\documentclass[runningheads]{llncs}

\usepackage{tabularx}

\usepackage[T1]{fontenc}

\usepackage{graphicx}

\usepackage{color}

\usepackage[utf8]{inputenc} % allow utf-8 input
\usepackage[T1]{fontenc}    % use 8-bit T1 fonts
\usepackage[pagebackref=true,breaklinks=true,colorlinks,bookmarks=false]{hyperref}

\usepackage{url}            % simple URL typesetting
\usepackage{booktabs}       % professional-quality tables
\usepackage{amsfonts}       % blackboard math symbols
\usepackage{nicefrac}       % compact symbols for 1/2, etc.
\usepackage{microtype}      % microtypography
\usepackage[table,dvipsnames]{xcolor}
\usepackage{wrapfig}
\usepackage{lipsum}
\usepackage{mwe}
\usepackage{amsmath,amsfonts}
\usepackage{algorithmic}
\usepackage{algorithm}
\usepackage{array}
\usepackage[caption=false,font=normalsize,labelfont=sf,textfont=sf]{subfig}
\usepackage{textcomp}
\usepackage{stfloats}
\usepackage{url}
\usepackage{verbatim}
\usepackage{graphicx}
\usepackage{multicol}
\usepackage{multirow}
\usepackage{enumitem}
\usepackage[textsize=tiny]{todonotes}
\usepackage{orcidlink}

\usepackage[utf8]{inputenc}
\usepackage{soul}
\usepackage{pifont}
\usepackage{amsmath}
\usepackage{amssymb}
\usepackage{bm}
\usepackage{color,soul}
\usepackage[dvipsnames]{xcolor}
\usepackage{gensymb}

\usepackage{xcolor}
\hypersetup{
    colorlinks,
    linkcolor={red!80!black},
    citecolor={blue!50!black},
    urlcolor={blue!80!black}
}

\newcommand{\mytilde}{\raise.17ex\hbox{$\scriptstyle\mathtt{\sim}$}}

\begin{document}

\title{Back to Supervision: Boosting Word Boundary Detection through Frame Classification}

\titlerunning{Boosting Word Boundary Detection through Frame Classification}

\author{
    Simone Carnemolla\inst{1}\orcidlink{0009-0005-8367-3933} \and
    Salvatore Calcagno\inst{1}\orcidlink{0000-0002-9763-213X} \and
    Simone Palazzo\inst{1}\orcidlink{0000-0002-2441-0982} \and
    Daniela Giordano\inst{1}\orcidlink{0000-0001-5135-1351}
}
\authorrunning{S. Carnemolla et al.}

\institute{
University of Catania, Department of Electrical Electronic and Computer Engineering, Via Santa Sofia, 95123 Catania, Italy\\
\email{simone.carnemolla@phd.unict.it}\\ 
}

\maketitle

\begin{abstract}
Speech segmentation at both word and phoneme levels is crucial for various speech processing tasks. It significantly aids in extracting meaningful units from an utterance, thus enabling the generation of discrete elements. In this work we propose a model-agnostic framework to perform word boundary detection in a supervised manner also employing a labels augmentation technique and an output-frame selection strategy. We trained and tested on the Buckeye dataset and only tested on TIMIT one, using state-of-the-art encoder models, including pre-trained solutions (Wav2Vec 2.0 and HuBERT), as well as convolutional and convolutional recurrent networks. Our method, with the HuBERT encoder, surpasses the performance of other state-of-the-art architectures, whether trained in supervised or self-supervised settings on the same datasets. Specifically, we achieved F-values of 0.8427 on the Buckeye dataset and 0.7436 on the TIMIT dataset, along with R-values of 0.8489 and 0.7807, respectively. These results establish a new state-of-the-art for both datasets. Beyond the immediate task, our approach offers a robust and efficient preprocessing method for future research in audio tokenization.

\end{abstract}

\section{Introduction}
Speech segmentation, from a psychological perspective, is the process by which our brain determines where a meaningful linguistic unit ends and the next begins in continuous speech~\cite{cc20ebb9eb49499fb35236b47ce12cbb}.

In machine learning before and in deep learning nowadays, this capability is not easily achievable due to the dense information that the audio data conveys and the prosodic features that each speaker has. Furthermore, building specialized datasets, with high-quality recordings and well-annotated data, requires considerable effort. More specifically, in speech processing, word and phoneme boundary detection are intended both as a preprocessing phase for other downstream tasks such as speaker diarization, keyword spotting, or automatic speech recognition, and as a tool to extract semantically meaningful information from audio.

 A correct boundary detection of words within a utterance would lead to a more accurate study of the prosody or emotional traits of speakers on a large scale, without necessary resorting to textual data alignment, that by their nature, lack important para-verbal information. Additionally, it would facilitate the discretization or tokenization of the elements that make up an utterance, a complex process given the variable length of speech units.

Due to the significant impact we believe this task may have on various downstream audio applications, we have opted for a supervised approach to maximize its performance, diverging from the self-supervised trend of the recent researches.

In this paper, we introduce a model-agnostic framework for word boundary detection. Our methodology integrates frame classification based on the BIO (begin, inside, outside) format with a label augmentation technique — to address the imbalance between \textit{begin} and \textit{inside}/\textit{outside} frames — and a frame-selection strategy for post-processing. We trained various state-of-the-art models on the Buckeye dataset \cite{pitt2005buckeye}, a widely recognized benchmark for this task. To further validate and check the generalization capability of our method we also tested it on TIMIT dataset \cite{garofolo1993timit}. 

Our results indicate that our method set a new state-of-the-art on both the Buckeye and TIMIT datasets, achieving F-values of 0.8427/0.7436, and R-values of 0.8489/0.7807, respectively. The code is publicly available on Github\footnote{\url{https://github.com/simonecarnemolla/Word-Segmenter}.}.

\section{Related work}
\label{section: Related_work}
Unlike phoneme boundary detection, which has a rich literature on supervised~\cite{keshet05_interspeech,mcauliffe17_interspeech,franke2016phoneme,kreuk2020phoneme}, self-supervised~\cite{strgar2023phoneme,wang17m_interspeech,kreuk20_interspeech}, and unsupervised~\cite{michel-etal-2017-blind,ALMPANIDIS200838,dusan06_interspeech,bhati17_interspeech} methods, the task of word boundary detection has been approached mainly from a self-supervised or unsupervised perspective.

Within supervised learning, most research focused its attention on probabilistic approaches and on the extraction of acoustic features making the preprocessing phase often long and complex. This is the case of Agarwal et al.~\cite{agarwal2010word} and Naganoor et al.~\cite{naganoor2016word}. The latter proposed a method that extracts rudimentary acoustic features and higher-order statistical features (HOS) . The same work inspired Shezi et al.~\cite{shezi2020word} for a word boundary detection task in IsiZulu language. 
In contrast, our approach focuses on utilizing raw audio data, sidestepping the need for extensive preprocessing steps. 

Other methods such as~\cite{mcauliffe17_interspeech} use word boundaries as a speech-text alignment system. However, our method takes a different route. We deliberately steer clear of relying on text label data, which sets our approach apart. This deliberate choice provides a significant advantage: our evaluation of performance is solely based on speech output. This independence from text data is a crucial feature of our method, rendering it particularly advantageous in situations where text data is scarce or unavailable. This characteristic ensures the robustness and applicability of our approach across various speech processing tasks, even in challenging data environments. 

\begin{figure}
\includegraphics[width=\textwidth]{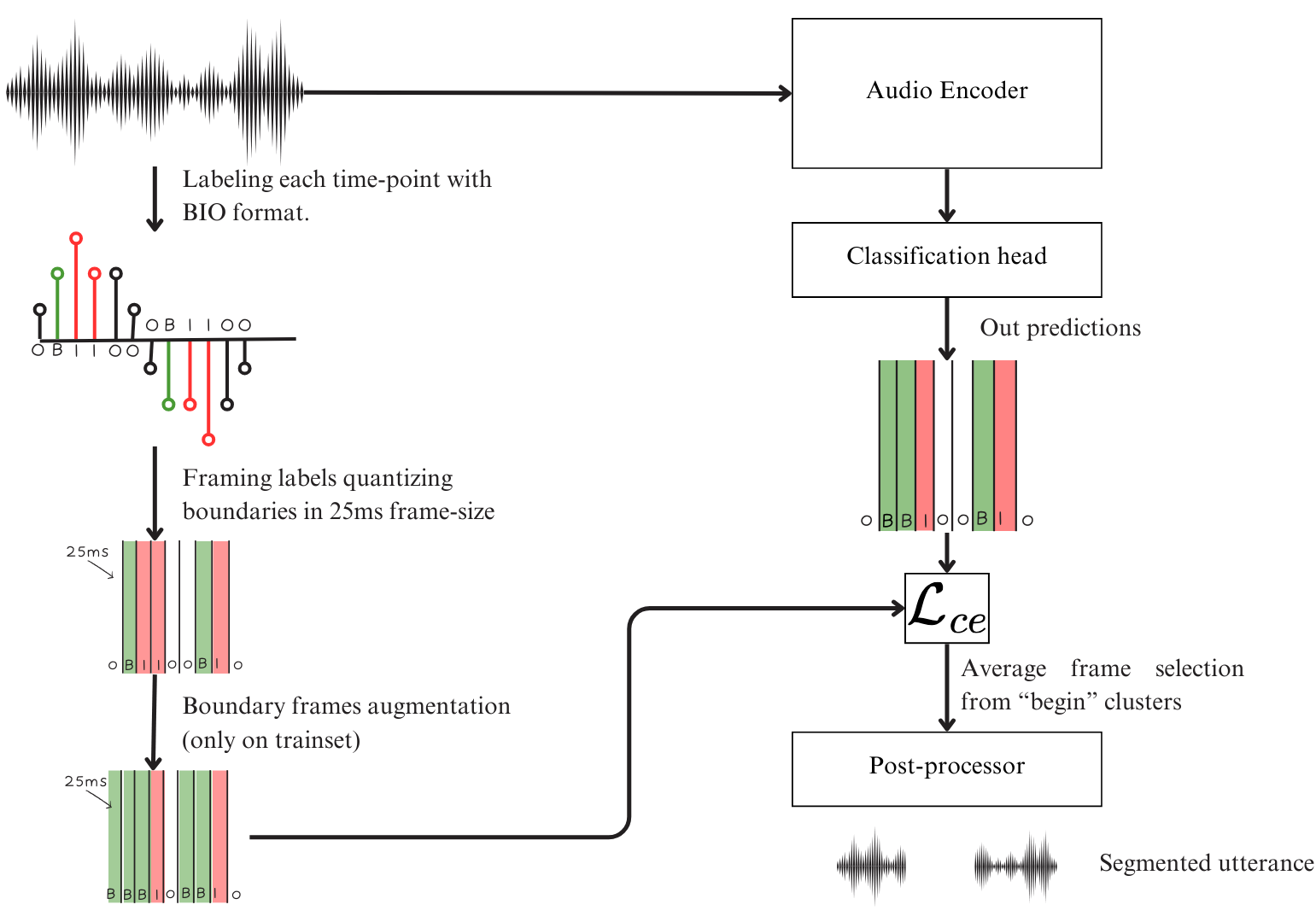}
\caption{Overview of our method} \label{fig1}
\end{figure}

Moving to self-supervised and unsupervised fields, some of the main contributions come from the works of Kamper et al.~\cite{kamper2017segmental,kamper2017embedded,kamper2016unsupervised}. In both~\cite{kamper2017segmental,kamper2016unsupervised}, a Bayesian model that segments unlabeled speech and clusters the segments into hypothesized word groupings is proposed. In a subsequent work~\cite{kamper2017embedded}, an embedded segmental KMeans (ES-KMeans) model is proposed to solve the Bayesian model's difficulty in scaling large speech corpora. Among the most interesting recent research advances are word boundary detection with vector quantized (VQ) neural networks~\cite{kamper21_interspeech}, a segmental contrastive predictive coding (SCPC) approach~\cite{bhati21_interspeech,bhati2022unsupervised}, and the use of temporal gradients as pseudo-labels to find boundaries~\cite{fuchs2023unsupervised}. In the first work~\cite{kamper21_interspeech}, the VQ neural networks – a vector-quantized variational autoencoder (VQ-VAE)~\cite{chorowski2019unsupervised} and a vector-quantized contrastive predictive coding (VQ-CPC)~\cite{Baevski2020} – are trained in a self-supervised way, segmenting speech into discrete units, assigning blocks of contiguous feature vectors to the same code. Then, dynamic programming (DP) is used to merge frames and to optimize a quadratic error with a length penalty term to encourage fewer but longer segments. Bhati et al.~\cite{bhati21_interspeech,bhati2022unsupervised} proposed a model that initially extracts frame-level representations and then identifies variable-length segments using a differentiable boundary detector. Finally, Fuchs et al.~\cite{fuchs2023unsupervised} extracted temporal gradients and observed that gradients with low magnitude effectively identify far-from-boundary regions. Building on this observation, they proposed GradSeg, a method where frames with gradient magnitudes below a preset threshold are assigned a positive label (indicating far-from-boundary words). This approach outperformed other unsupervised methods. Furthermore, Fuchs et al. trained with a supervised approach to compare the impact of supervision on the results. We employed their supervised results as a state-of-the-art benchmark, as we surprisingly found no other recent methods to compare with. We also reproduced the experiments using our data distribution with their unsupervised method and their earlier work \cite{fuchs22_interspeech}, reporting the scores. This was done not to compare their results with ours, but to validate the assumption that a supervised method can significantly enhance the applicability of word segmentation.

\section{Method}
\subsection{Overview}
An overview of our method is shown in Figure \ref{fig1}. Initially, the raw audio is labeled using a BIO format. Following framing, we apply a label augmentation technique to better handle the significant imbalance between beginning and inside/outside indices. An encoder architecture is then trained in a supervised setting for the frame classification task. Finally, a frame selection strategy is employed to post-process predictions and segment the input utterance. Each component of the method is described in detail in the following subsections.
\subsection{Problem Formulation}
\label{problem_formulation}
Let $\mathbf{x}_i=(x_{i,1},x_{i,2},\ldots,x_{i,T})$ be a single utterance where $x_t$ represents the amplitude of the signal at time $t$, with $0 \leq t \leq T$, considering a sampling frequency $s$. Let also denote $\mathbf{y}_i=(y_{i,1},y_{i,2},\ldots,y_{i,m})$ the corresponding sequence of framed labels. The generic value $y_{i,j}$ refers to a frame within $\mathbf{x}_i$. All frames have the same fixed time-length (more details in Section \ref{labels_processing}). In particular, $y_{i,j} \in \{0,1,2\}$ indicates if the corresponding frame in $\mathbf{x}_i$ marks the start, is positioned inside, or lies outside a single word. Our objective is to accurately detect the word boundaries within the utterance $\mathbf{x}_i$ by predicting the correct sequence of framed labels $\mathbf{y}_i$. Once the boundary frames have been identified, we aim to identify the exact time-point in the audio signal where words begin.

We adopted a multi-class cross-entropy as a loss function, defined as follows:
\begin{equation}
    \mathcal{L}(\mathbf{y},\mathbf{\hat{y}}) = -\frac{1}{N}\sum_{i=1}^{N}\sum_{j=1}^{m}y_{i,j}\cdot\log(\hat{y}_{i,j}),
    \label{equation:eq1}
\end{equation}

\noindent Where $N$ is the number of instances in the batch, $y_{i,j}$ represents the label of the $j$-th frame of the $i$-th dataset element, and $\hat{y}_{i,j}$ represents the model’s prediction for the same frame.
\subsection{Labels Processing}
\label{labels_processing}
As showed in Fig.\ref{fig1}, we extracted all the utterances and the start-end boundaries of each word per utterance. We maintained the sample rate of the audio files at 16KHz without resampling. Subsequently, we created pre-labels with the same shape of the input waveforms, assigning to each time-point a value depending on if it was positioned inside or outside the boundaries or if it was a start index. This step was useful to investigate the distribution and the average duration of the words within the utterances. The final labels were obtained, framing the pre-labels along the temporal axis with a 25 ms frame duration. The total number of frames $m$ in $\mathbf{y_i}$ is given by:

\begin{equation}
    m = \frac{{T}}{{25 \cdot \textit{s}}}
\end{equation}

\noindent where $T$ is the number of time-points of the input sequence $\mathbf{x_i}$ and $s$ is its sampling frequency.

\noindent To address the significant imbalance of labels (i.e., begin, inside, and outside annotations) during training, we also considered the frames adjacent to the ground truth as \textit{begin}. 
Specifically, we selected one frame to the left and one frame to the right of the actual start. We did not apply this augmentation during inference. We observed that this approach, combined with the frame selection strategy described in Section \ref{post-processing}, considerably improve the final scores. More details about the interplay of labels augmentation and frame selection are described in Section \ref{label_aug_frame_sel}.

\subsection{Model Architecture}
\label{encoder} Given an input utterance \(\mathbf{x_i}\) as described in Section \ref{problem_formulation}, the waveform passes through an audio encoder \(A_{enc}\), resulting in a hidden representation \(\mathbf{z_i} \in \mathbb{R}^{n \times d}\), where \(n\) is the number of frames produced by the encoder and \(d\) is the dimension of the hidden state.

\begin{equation}
    \mathbf{z_i} = A_{enc}(\mathbf{x_i})
\end{equation}

\noindent Then, a linear  layer $A_{lp}$ projects $\mathbf{z_i}$ into $\mathbf{e_i} \in \mathbb{R}^{d \times m}$, in order to match the dimensions of $\mathbf{y_i}$.

\begin{equation}
    \mathbf{e_i} = A_{lp}(\mathbf{z_i}^{\top})
\end{equation}

\noindent Finally, a linear classification head \(A_{lc}\) is applied to obtain the predictions \(\mathbf{c_i} \in \mathbb{R}^{m \times p}\), where \(p\) represents the output probabilities for each class.

\begin{equation}
    \mathbf{c_i} = A_{lc}(\mathbf{e_i}^{\top})
\end{equation}

\subsection{Post-processing}
\label{post-processing}
We extract word boundaries and post-process the audio input to return a sequence of variable-length segments, each corresponding to a word. Given the augmentation strategy outlined in Section \ref{labels_processing}, the model tends to over-segment during inference, resulting in clusters of predicted \textit{begin} frames. To mitigate this behaviour we select the average frame as predicted one discarding its neighbors. During the training stage no tolerance was applied, while at inference time, to check the correctness of the boundaries and compute the metrics, we consider a tolerance of 40 ms as \cite{fuchs2023unsupervised}. Differently from this work we compare predicted boundaries with the time-point level ground truth instead of the frame quantized one, resulting in a more accurate evaluation. Testing on TIMIT dataset we applied a tolerance of 20 ms to align our scores with the ones got by \cite{bhati2022unsupervised}.

\begin{table}

    \centering

        \setlength{\tabcolsep}{15pt} % Default value: 6pt
        \renewcommand{\arraystretch}{1.2}
        \caption{Architectures of employed models. We report convolutional layer details (channels, kernel size, stride), hidden size for recurrent model. For Wav2Vec, and HuBERT please refer to 
        \cite{baevski2020wav2vec}\cite{hsu2021hubert}}.
    
    \label{tab:models_configuration}
    
    \begin{tabular}{l|cc} 

         \toprule
         & \textbf{CNN} &  \textbf{CRNN}\\

         \midrule
         \rowcolor{gray!10}
         \# Convolutional channels &  16, 32, 64, 128 &  16, 32, 64, 128\\ 
         Kernel size &  11, 3, 3, 3&  11, 3, 3, 3\\ 
         \rowcolor{gray!10}
         Stride&  5, 2, 2, 2&  5, 2, 2, 2\\ 
        Hidden size& -&80, 40\\
        \bottomrule
    \end{tabular}
\end{table}

\section{Experimental setup}
\subsection{Datasets}
Buckeye \cite{pitt2005buckeye} is a spontaneous speech corpus containing recordings of 40 talkers from central Ohio interviewed for about one hour. The group of people is stratified on age and gender and each recording is sampled at 16Khz and annotated to phoneme and word level. For our purpose we used the word annotations only. The whole corpus counts about 307,000 words.

TIMIT \cite{garofolo1993timit} is a speech corpus used for acoustic-phonetic studies, but it was also frequently employed for phoneme and word boundary detection tasks \cite{kreuk2020phoneme,kreuk20_interspeech,bhati21_interspeech,bhati2022unsupervised}. It comprises recordings from 630 speakers, including 438 males and 138 females. Each speaker recorded ten utterances, resulting in a total of 6300 speech samples. These utterances are phonetically and lexically annotated, with indices marking the start and end of phonemes and words. For our study, we utilized only the test set, which includes 1680 samples.

On Buckeye, considering the length of each audio recording, we truncate them and split in train, validation and test with a similar strategy employed by \cite{fuchs2023unsupervised}. We considered to use this way also to facilitate the comparison of results and the reproduction of the experiments.

\subsection{Audio Pre-processing}
The only manipulations applied on audio were standardization and padding. Since waveforms were with variable length, the standardization was done by calculating the global weighted mean and global weighted standard deviation of the train set. The same values were then applied to the validation and test set. The padding was applied based on the longest waveform corresponding to a duration of approximately 9 seconds.
\subsection{Encoders}
Our framework is model-agnostic, meaning it can be applied regardless of the encoder architecture. However, in order to evaluate the method we chose several well-known architectures. Specifically, we selected a one-dimensional CNN and CRNN as from-scratch architectures, while we employed Hubert and Wave2Vec as pretrained models.
The CNN takes raw audio as input and consists of four convolutional layers followed by batch normalization, ReLU activation, and max pooling, with the number of filters doubling at each layer. Its output is transposed and passed to a fully connected layer that produces a representation $\mathbf{z} \in \mathbb{R}^{d \times m}$, where $d$ are the convolutional features and $m$ the number of framed labels. Finally, the resulting vector is transposed again and passed to a linear classification layer.  
We chose the CRNN because it represents the state-of-the-art for various previous works in audio segmentation and classification~\cite{venkatesh2021artificially,venkatesh2021investigating,app12073293,salamon2017scaper}. The network retains the same configurations as the CNN model for the convolutional layers, but without batch normalization. Additionally, it introduces two bidirectional Gated Recurrent Unit (B-GRU) layers with a similar configuration to ~\cite{venkatesh2021artificially,venkatesh2021investigating}. A final linear projection and a linear classification layer are applied. CNN and CRNN configurations are showed in Tab. \ref{tab:models_configuration}.

We decided also to include the two main pretrained models at the state-of-the-art for several downstream tasks: HuBERT$_{Large}$ \cite{hsu2021hubert} and Wav2Vec2.0$_{Base}$ \cite{baevski2020wav2vec}. We kept both encoders frozen while fine-tuning the final linear layers, which were followed by layer normalization.
\subsection{Training Procedure}
\label{training_inference}
All the models were trained on a NVIDIA RTX A6000. The average training time of our best fine-tuned model (i.e. HuBERT encoder) is around one hour. The inference time for the whole Buckeye test set is 26 seconds.
We tuned learning rate and batch size hyperparameters with grid search on the validation set, choosing at the end $10^{-3}$ as learning rate and 32 as batch size. We also applied an early stopping with a patience of 10 epochs if no improvement occurred on the best validation R-value. We did not use time error tolerance for the boundaries detection during the training phase. More details are available in the supplementary materials.

\subsection{Metrics}
The set of metrics, as defined in \cite{aversano2001new,petek1996robust,ajmera2004robust,rasanen2009improved}, is composed by Precision, Recall, F-value, Over Segmentation (OS), and R-value.

\noindent\textbf{Precision} (PRC) and \textbf{Recall} (RCL) were employed as described by \cite{rasanen2009improved} and expressed in \ref{equation:Precision, Recall}. In the equation $\textit{N}_{\textit{hit}}$ represents the boundaries correctly detected, while $\textit{N}_{\textit{ref}}$ stands for the total number of boundaries in the reference.

\begin{equation}
\textit{PRC} = \frac{\textit{N}_{\textit{hit}}}{\textit{N}_{\textit{f}}} , \textit{RCL} = \frac{\textit{N}_{\textit{hit}}}{\textit{N}_{\textit{ref}}}
\label{equation:Precision, Recall}
\end{equation}

\noindent\textbf{OS} \cite{petek1996robust} is the over segmentation rate and is given by the ratio of the total number of detected boundaries $\textit{N}_{\textit{f}}$ over the total number of boundaries $\textit{N}_{\textit{ref}}$ in the reference. Then the result is subtracted by one (Eq.\ref{equation:os}).  

\begin{equation}
    \textit{OS}=\frac{\textit{N}_{\textit{f}}}{\textit{N}_{\textit{ref}}}-1
    \label{equation:os}
\end{equation}

\noindent\textbf{F-value} \cite{ajmera2004robust} is the harmonic average of PRC and RCL.

\begin{equation}
\textit{F-value} = \frac{2 \cdot \textit{PRC} \cdot \textit{RCL}}{\textit{PRC} + \textit{RCL}}
\label{equation:f_value}
\end{equation}

\noindent\textbf{R-value} \cite{rasanen2009improved} is another composed metric derived from OS and Recall and is defined as a trade-off of these two metrics, being the balance between Recall and OS a suitable operating point for audio segmentation. R-value can be expressed as follow:

\begin{equation}
r_1 = \sqrt{(1 - \textit{RCL})^2 + \textit{OS}^2},  r_2 = \frac{-\textit{OS} + \textit{RCL} - 1}{ \sqrt{2}}
\end{equation}

\begin{equation}
\textit{R-value} = 1 - \frac{\lvert r_1 \rvert + \lvert r_2 \rvert}{2}
\end{equation}

\noindent We decided to use R-value as main quality metric to evaluate the models, i.e we saved the weights of models at the best validation R-value. 
\subsection{Experiments}
To evaluate our method, we trained from scratch the CNN and CRNN architectures and fine-tuned HuBERT and Wav2Vec keeping both encoders frozen. For a comparative analysis we reproduced the training procedures of GradSeg \cite{fuchs2023unsupervised} and DSegKNN \cite{fuchs22_interspeech}. All the architectures were tested on our Buckeye test set.
To assess the generalization capability of our method, we also tested the above methods on the TIMIT dataset.
Finally, to underline the need for label augmentation and frame selection, we assessed the performance of our method with and without the application of the two strategies.

 %describe 

%\begin{figure}[htp]

%    \centering
%    \includegraphics[width=0.45\textwidth]{img/correctmatch_5.png}
%   \includegraphics[width=0.45\textwidth]{img/correctmatch_4.png}\\

%    \includegraphics[width=0.45\textwidth]{img/correctmatch_2.png}
%    \includegraphics[width=0.45\textwidth]{img/correctmatch_3.png}\\

%    \includegraphics[width=0.45\textwidth]{img/undersegmentation_0.png}
%    \includegraphics[width=0.45\textwidth]{img/undersegmentation_2.png}\\

%    \includegraphics[width=0.45\textwidth]{img/undersegmentation_3.png}
%   \includegraphics[width=0.45\textwidth]{img/oversegmentation_1.png}\\
    
%    \caption{Segmentation comparison example between true (green solid line) and predicted (red dashed line) boundaries on Buckeye test set. The four pictures on top show a perfect match of detected boundaries. The next three pictures report an under segmentation, while the last one an over segmentation.}
%\label{fig:fig2}
    
%\end{figure}

\begin{figure}[htp]
    \centering
    
    \begin{tabular}{cc}
        \multicolumn{2}{c}
        {\textbf{Correct matches}} \\
        \includegraphics[width=0.45\textwidth]{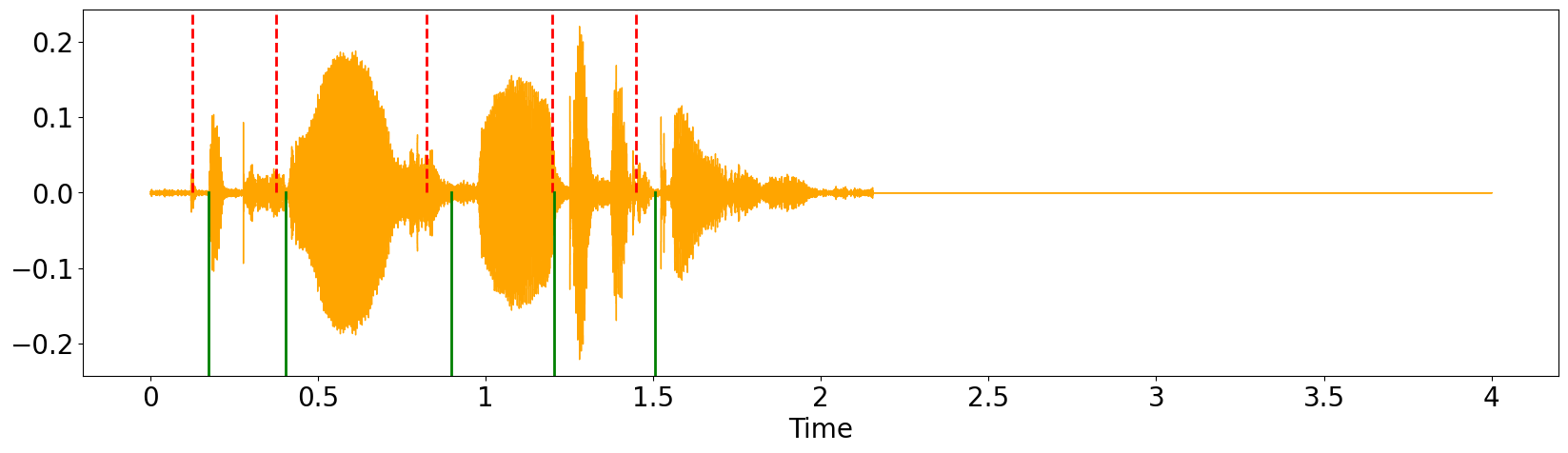} &
        \includegraphics[width=0.45\textwidth]{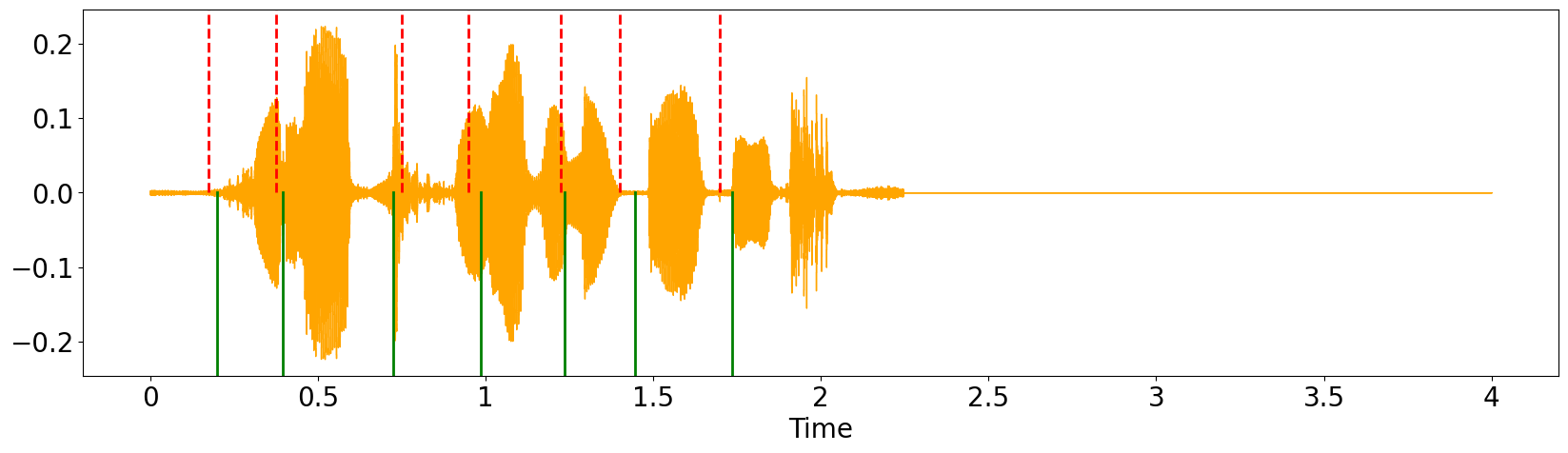} \\
        \includegraphics[width=0.45\textwidth]{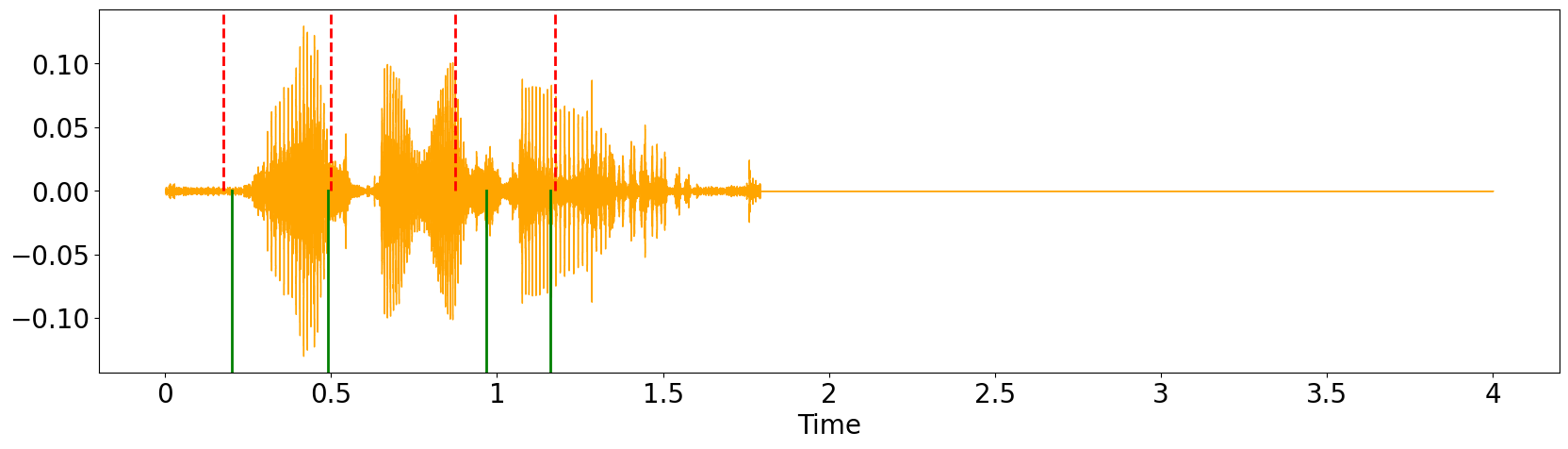} &
        \includegraphics[width=0.45\textwidth]{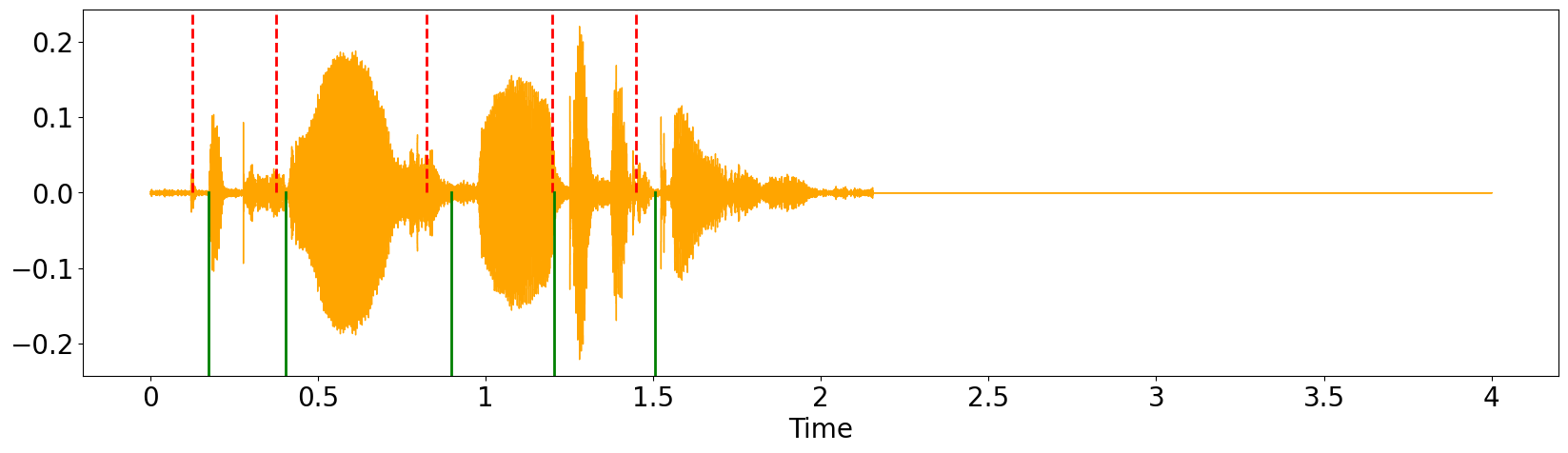} \\
        \multicolumn{2}{c}{\textbf{Under/Over segmentation}} \\
        \includegraphics[width=0.45\textwidth]{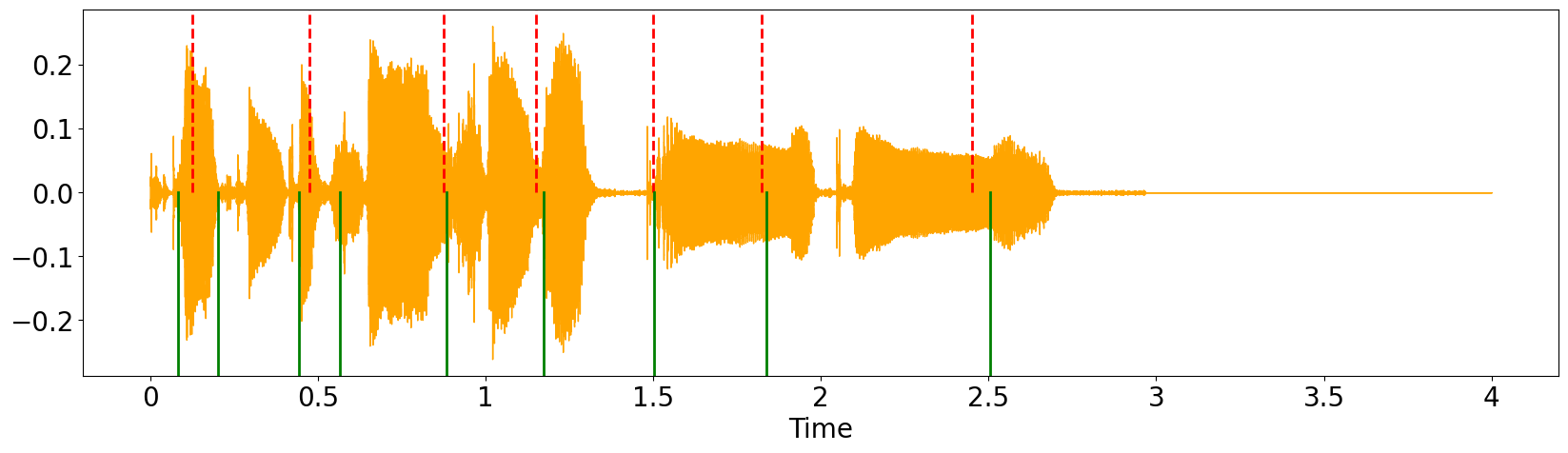} &
        \includegraphics[width=0.45\textwidth]{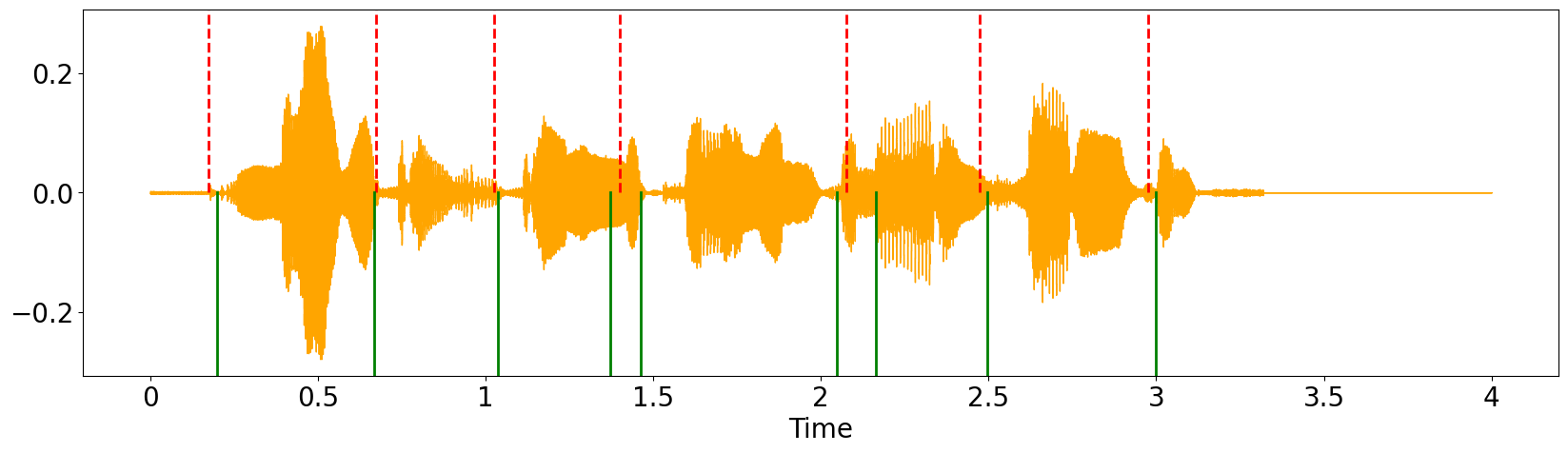} \\
        \includegraphics[width=0.45\textwidth]{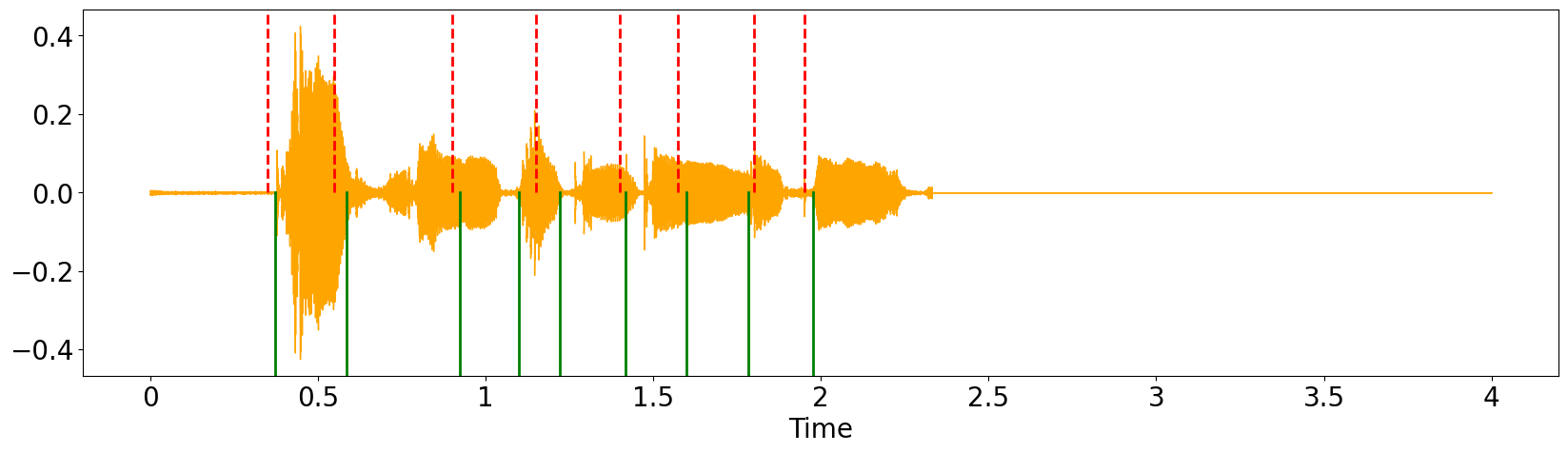} &
        \includegraphics[width=0.45\textwidth]{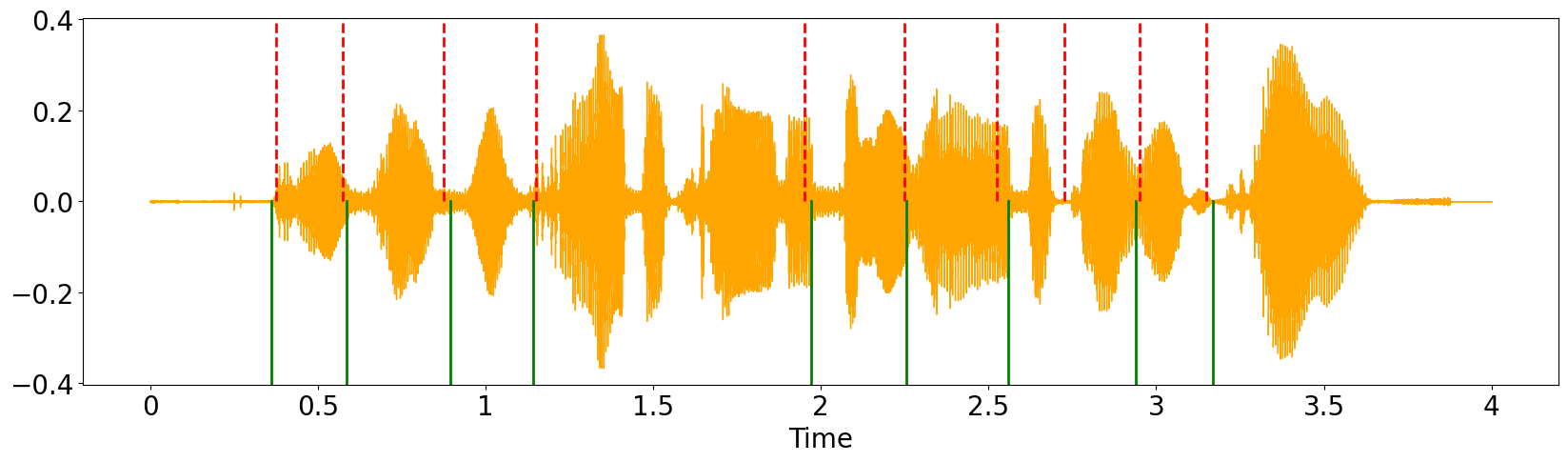}
    \end{tabular}
    
    \caption{Segmentation comparison example between true (green solid line) and predicted (red dashed line) boundaries on Buckeye test set. The four pictures on top show a correct match of detected boundaries. The second group report  three under segmentation scenarios and an over segmentation one.}
    \label{fig:fig2}
    
\end{figure}

\begin{table*}[htb!]

    \centering
    \setlength{\tabcolsep}{8pt} % Default value: 6pt
    \renewcommand{\arraystretch}{1.2}

    \caption{Comparison of not-pretrained/pretrained supervised models and unsupervised ones on Buckeye test set. Tolerance was set to 40ms.}
    
    \label{tab:Scores}
    
    \begin{tabular}{l|ccccc}

    \toprule
    \multicolumn{6}{c}{\textbf{Unsupervised models}}\\

    \midrule
    \textbf{Model}& \textbf{Precision}& \textbf{Recall}& \textbf{F-value}& \textbf{OS} &\textbf{R-value}\\

    \midrule
    \rowcolor{gray!10}
    DSegKNN \cite{fuchs22_interspeech}& 0.3115& 0.3226& 0.3169& 0.0355&0.4087\\
    
    GradSeg \cite{fuchs2023unsupervised}& 0.4444& 0.4356& 0.4399& \textbf{-0.0197}&0.5251\\

    \toprule
    \multicolumn{6}{c}{\textbf{Supervised models}}\\

    \midrule
    \textbf{Model}& \textbf{Precision}& \textbf{Recall}& \textbf{F-value}& \textbf{OS} &\textbf{R-value}\\

    \midrule
    \rowcolor{gray!10}
    CNN&  0.3842&  0.3604&  0.3708&  -0.0575&0.4694\\
    
    CRNN&  0.4112&  0.3711&  0.3896&  -0.0972&0.4923\\
    
    \rowcolor{gray!10}
    GradSeg \cite{fuchs2023unsupervised}&  -&  -&  0.5960&  -&-\\
    
    Wav2Vec$_\textit{Base}$ \cite{baevski2020wav2vec}&  0.6556&  0.4736&  0.5494&  -0.2766&0.6139\\
    \rowcolor{gray!10}HuBERT$_\textit{Large}$ \cite{hsu2021hubert}&  \textbf{0.8999}&  \textbf{0.7928}&  \textbf{0.8427}&  -0.1187&\textbf{0.8489}\\ 
    \bottomrule
    \end{tabular}
\end{table*}

\section{Results}
In the subsequent sections, we present the obtained results. Our method is benchmarked against a single supervised reference for Buckeye, while for TIMIT no supervised methods have been found. Finally, additional results without comparisons on NTIMIT \cite{jankowski1990ntimit} are reported in the supplementary materials. As discussed in Section \ref{section: Related_work}, the prevailing trend in recent research on word boundary detection involves self-supervised or unsupervised approaches making complex the possibility of comparison with those who want to tackle this task in a supervised way. On the other hand, a direct comparison with our supervised method may not be entirely fair. Nevertheless, we opted to include results from unsupervised methods, not for direct comparison with our approach, but to underscore the potential impact of a supervised approach to word boundary detection in advancing applications within speech technology. 

It is worth noting that certain supervised methods leverage word boundary detection for aligning speech to text transcriptions. For this task, metrics such as Word Error Rate (WER) are employed to assess the models and their efficacy. Given the disparate objectives of these metrics compared to ours, a direct correlation might be deemed unfair.

\subsection{Models from scratch}
As shown in Table \ref{tab:Scores}, the CNN model achieved lower results compared to its unsupervised counterpart. The CRNN model performed better than the CNN in most metrics except for the OS metric, yet it still lower than \cite{fuchs2023unsupervised}.

It is important to note that while the scores of both CNN and CRNN models are lower than those reported by \cite{fuchs2023unsupervised}, the latter utilized Wav2Vec pretrained as an encoder.

Table \ref{tab:Scores_TIMIT} presents the scores obtained on the TIMIT dataset \cite{garofolo1993timit}. As anticipated, both models demonstrated a general decline in performance compared to the results on the Buckeye dataset. Additionally, their scores are lower than those reported by \cite{bhati2022unsupervised}. However, it is important to consider that \cite{bhati2022unsupervised} was specifically trained on the TIMIT dataset.

\begin{table*}[htb!]
    \centering
    \setlength{\tabcolsep}{10pt} % Default value: 6pt
    \renewcommand{\arraystretch}{1.2}
    \caption{Comparison of not-pretrained/pretrained supervised models and unsupervised one on TIMIT test set. Tolerance was set to 20 ms.}
    \label{tab:Scores_TIMIT}
    \begin{tabular}{l|ccccc}
    
    \toprule
    \multicolumn{6}{c}{\textbf{Unsupervised models}}\\

    \midrule
     \textbf{Model}& \textbf{Precision}& \textbf{Recall}& \textbf{F-value}& \textbf{OS} &\textbf{R-value}\\
     
    \rowcolor{gray!10}
    \midrule
    SCPC \cite{bhati2022unsupervised}& 0.2895& 0.2302& 0.2558& -&0.4003\\

    \toprule
    \multicolumn{6}{c}{\textbf{Supervised models}}\\

    \midrule
     \textbf{Model}& \textbf{Precision}& \textbf{Recall}& \textbf{F-value}& \textbf{OS} &\textbf{R-value}\\
     
    \rowcolor{gray!10}
    CNN&  0.2490&  0.1697&  0.2008&  -0.3169&0.3731\\
    
    CRNN&  0.2610&  0.2247&  0.2411&  -0.1377&0.3794\\
    
    \rowcolor{gray!10}Wav2Vec$_\textit{Base}$ \cite{baevski2020wav2vec} &  0.4433&  0.3538&  0.3930&  -0.2005&0.5032\\
    
    HuBERT$_\textit{Large}$ \cite{hsu2021hubert} &  \textbf{0.7566}&  \textbf{0.7314}&  \textbf{0.7436}&  \textbf{-0.032}&\textbf{0.7807}\\
         \bottomrule
    \end{tabular}
\end{table*}

\subsection{Pretrained models}
HuBERT outperforms all models except in the over-segmentation metric, where GradSeg \cite{fuchs2023unsupervised} in its unsupervised version achieved the value closest to zero. Additionally, we observed that both Wav2Vec and HuBERT exhibit higher Precision than Recall. This behavior is likely due to the frame selection strategy described in Section \ref{training_inference}. Specifically, by selecting only the average boundary from the boundary clusters, we penalize instances where adjacent words (true positives) occur. Although we tested other frame selection strategies (see Section \ref{label_aug_frame_sel}) to improve the recall metric, we ultimately favored a more precise model over a more sensitive one.

In the GradSeg article \cite{fuchs2023unsupervised}, the authors reported only the F-value metric for the supervised method, probably because it was not the main focus of their work. Nonetheless, this metric allowed us to compare our models with another benchmark, as we encountered difficulty finding recent supervised methods for comparison. Fig.\ref{fig:fig2} shows some predictions done on Buckeye utterances by HuBERT model compared with the ground truth boundaries. The model tends to an under-segmentation, however the predicted boundaries are frequently close to the real ones.

Also on TIMIT dataset (Tab. \ref{tab:Scores_TIMIT})  HuBERT outperforms the other models.

This section demonstrates that utilizing pretrained encoders for supervised training, along with other strategies like labels augmentation and output-frame selection, can significantly enhance the quality of word segmentation. This improvement is evident not only on the specific dataset used to train the models, but also on other speech datasets (such as TIMIT) , achieving even better results than unsupervised methods trained on them \cite{bhati2022unsupervised}.

\begin{table*}[htb!]
    \centering
    \renewcommand{\arraystretch}{1.2}
    \setlength{\tabcolsep}{5pt} % Default value: 6pt
    \caption{Ablation study on labels augmentation and frame selection.}
    \footnotesize{
    \begin{tabular}{ll|ccccc}
        \toprule
        & \textbf{Model}&  \textbf{Precision}& \textbf{Recall}&\textbf{F-value}&  \textbf{OS}&\textbf{R-value} \\
        \midrule
        \rowcolor{gray!10}
        & Wav2Vec$_{Base}$&  0.8805& 0.0460&0.0874&  -0.9475&0.3254\\ 
        & \hspace{0.1 cm} {$\hookrightarrow$Label Augmentation}&  0.3013& 0.6585&0.4131&  1.1889&-0.1604\\
        \rowcolor{gray!10}
        & \hspace{0.3 cm} {$\hookrightarrow$Frame selection}&  \textbf{0.6556}& \textbf{0.4736}&\textbf{0.5494}& \textbf{-0.2766}&\textbf{0.6139}\\ 
        \midrule       
        & HuBERT$_{Large}$&  0.9302& 0.4403&0.5971&  -0.5327&0.6032\\
        \rowcolor{gray!10}
        & \hspace{0.1 cm} {$\hookrightarrow$Label Augmentation} &  0.4840& 0.9161&0.6332& 0.8942&0.2049\\
 & \hspace{0.3 cm} {$\hookrightarrow$Frame selection}&  \textbf{0.8999}& \textbf{0.7928}&\textbf{0.8427}& \textbf{-0.1187}&\textbf{0.8489}\\

    \bottomrule
    \end{tabular}}
    \label{tab:labels_aug_fram_sel}
    
\end{table*}

\subsection{Effect of labels augmentation and frame selection}
\label{label_aug_frame_sel}
As shown in Table \ref{tab:labels_aug_fram_sel}, applying only labels augmentation considerably worsens the scores. This is because increasing the number of boundaries enhances the model's sensitivity, but it also leads to higher over-segmentation, significantly reducing precision and affecting both the F-value and R-value.

\begin{figure}[htp]
    \centering
    \begin{minipage}[t]{0.45\textwidth}
        \centering
        \includegraphics[width=\textwidth]{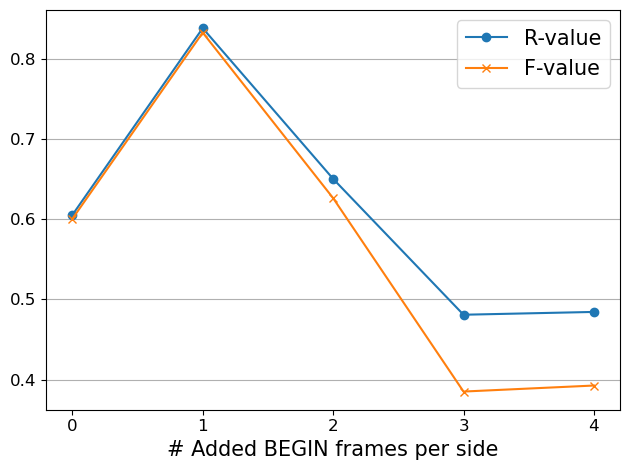}
        \caption{Comparison of R-value and F-value for the Buckeye validation set based on different window sizes for label augmentation. Each data point represents the number of frames labeled as \textit{begin} to the left and right of the ground truth. The results are computed employing the HuBERT encoder.}
        \label{fig3}
    \end{minipage}\hfill
    \begin{minipage}[t]{0.45\textwidth}
        \centering
        \includegraphics[width=\textwidth]{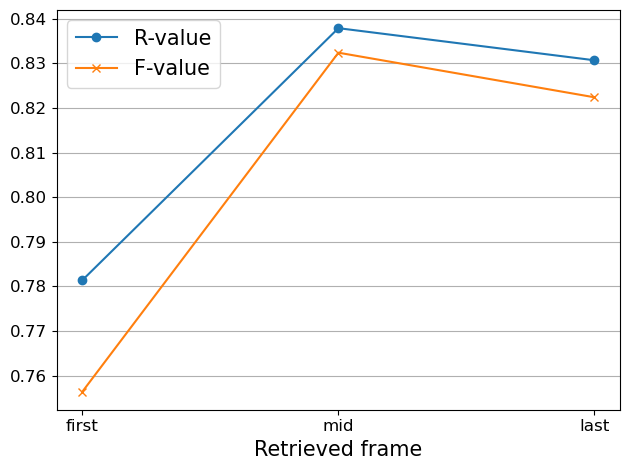}
        \caption{Comparison of R-value and F-value scores for the Buckeye validation set based on different frame selection strategies. The first approach retrieves the initial \textit{begin} frame from the \textit{begin} cluster, the second approach selects the middle frame, and the third approach picks the final frame. The results are computed employing the HuBERT encoder.}
        \label{fig4}
    \end{minipage}
\end{figure}

When the frame selection strategy is applied, we mitigate the over-segmentation issue, decreasing recall scores but improving overall performance.  Different experiments were conducted with "begin" clusters during labels augmentation as reported in Fig. \ref{fig3}. Ultimately, we chose to label as "begin" one frame to the left and one frame to the right of the actual boundary because this setting yielded the best performance.

On the frame selection side (Fig. \ref{fig4}), we tested three approaches: selecting the first frame, the last frame and the mid one. However, the best choice in terms of scores was to extract the mid-frame for each \textit{begin} cluster.

\subsection{Discussion}
As demonstrated in previous sections,  in Table \ref{tab:Scores} and in Table \ref{tab:Scores_TIMIT}, employing a supervised approach centered on frame classification significantly enhances the performance of word boundary detection (WBD). Notably, it's not solely the choice of approach (frame classification) that influences the outcomes, but also the methodology we employ in handling labels imbalance during training and frame selection during inference as showed in Tab. \ref{tab:labels_aug_fram_sel} and discussed in Section \ref{label_aug_frame_sel}, ensuring anyway that them don't alter the inherent nature of the data and maintains a streamlined preprocessing pipeline.
It's crucial to acknowledge the significance we attribute to the WBD task. In contrast to the self-supervised methods outlined in Section \ref{section: Related_work}, which strive to generate directly meaningful word-level audio latent representations, we interpret this task as an initial step to provide support and input for self-supervised models with discrete units, akin to tokens in text. In light of this intent, based on performance, we can state that the self-supervised model are not ready yet and probably this is not neither their goal.
With this work, we also aim to stimulate the audio community to explore other supervised methods for word boundary detection. We strongly believe that this approach could significantly boost the performance of this task and consequently enhance the development of recent audio application trends, such as speech-to-speech conversational models and real-time translators. 

\section{Conclusion}
In this work we propose a robust and computationally light preprocessing approach for word boundary detection and evaluated its efficacy compared to other supervised and unsupervised methods, by using pre-trained and from scratch solutions. Our future work will leverage extracted words to build a tokenization-like method, thus enabling the variable-length discrete units to retain important para-verbal and prosodical features and paving the way to stronger self-supervised models and spoken dialogue systems.

\subsubsection{Acknowledgements}
Simone Carnemolla and Salvatore Calcagno acknowledge financial support from: PNRR MUR project PE0000013-FAIR.

\bibliographystyle{splncs04}{
\bibliography{biblio}
}

\end{document}